# Exploring Tokenization Strategies and Vocabulary Sizes for Enhanced Arabic Language Models

Mohamed Taher Alrefaie[1], Nour Eldin Morsy[2], Nada Samir[2]

[1] Egypt University of Informatics, Knowledge City, New Administrative Capital, Egypt.
[2] AAST College of Artificial Intelligence, Alameen City, Egypt.

## Abstract

This paper presents a comprehensive examination of the impact of tokenization strategies and vocabulary sizes on the performance of Arabic language models in downstream natural language processing tasks. Our investigation focused on the effectiveness of four tokenizers across various tasks, including News Classification, Hate Speech Detection, Sentiment Analysis, and Natural Language Inference. Leveraging a diverse set of vocabulary sizes, we scrutinize the intricate interplay between tokenization approaches and model performance. The results reveal that Byte Pair Encoding (BPE) with Farasa outperforms other strategies in multiple tasks, underscoring the significance of morphological analysis in capturing the Arabic's complex nature. However, challenges arise in sentiment analysis, where dialect-specific segmentation issues impact model's efficiency. Computational efficiency analysis demonstrates the stability of BPE with Farasa, suggesting its practical viability. Contrary to conventional assumptions, our study uncovers limited impacts of vocabulary size on model performance while keeping the model size unchanged. This challenges the common understanding among the relationship between vocabulary, model size, and downstream tasks; emphasizing the need for the study of models' size and their corresponding vocabulary size to generalize across domains and mitigate biases, particularly in dialect-based datasets. Our findings suggest the need for a tokenization strategies to address dialect challenges, enhancing model robustness across diverse linguistic contexts, and expanding datasets to encompass the rich dialect-based Arabic.

# 1. Introduction

Deep language models have become the most important language modelling based on the Transformer architecture (Vaswani et al., 2017). This method, exemplified by BERT (Devlin et al., 2019; Liu et al., 2023) and its variations have created highly versatile language models. Subsequently, these models are finetuned on specific downstream tasks such as Sentiment Analysis and Named Entity Recognition to provide accurate classification and modelling. The tokenization step is the first step to transform text into vectors that can be fed into large language models. Tokenization utilizes efficient algorithms that split input text into smaller units, ensuring effective processing of out-of-vocabulary words. This approach enhances the models' comprehension of text semantics by employing tokens representing basic semantic units (Ali et al., 2023).

Over the past few years, the landscape of tokenization has undergone significant evolution, progressing from word-level tokenization with embeddings like Word2Vec and GloVe to n-gram embeddings and eventually to subword tokenization algorithms. Notably, subword tokenization, facilitated by BPE, WordPiece, and unigram language modelling, has gained prominence in influential models such as BERT, RoBERTA, and GPT-2, reflecting a paradigm shift in Large Language Models (LLMs) (Devlin et al., 2019; Kudo, 2018; Mikolov et al., 2013; Pennington et al., 2014; Sennrich et al., 2016a; Wu et al., 2016).

In the realm of Arabic Natural Language Processing (NLP), subword tokenization stands as a prevalent choice, evident in the adoption of WordPiece by HULMonA (ElJundi et al., 2019) and SentencePiece by AraBERT (Antoun et al., 2020). These models showcase the efficacy of subword tokenization, especially when integrated with FARASA morphological segmentation (Abdelali et al., 2016), across language modelling and text classification tasks, achieving advanced results.

Tokenization sensitivity in Arabic text processing emerges as a critical consideration, as demonstrated in studies assessing the impact on machine translation and BERT model fine-tuning. Oudah et al., (2019) argue that the choice of tokenization scheme is contingent upon both the data and the specific model employed for training, emphasizing the need for tailored approaches.

A series of studies have explored the impact of tokenization on the performance of large language models in morphologically rich languages. Toraman et al., (2022) found that a morphological-level tokenizer can outperform de facto tokenizers, with a larger vocabulary size improving performance. This is consistent with the findings of Choe et al., (2019), who demonstrated that tokenizer-free language models can achieve competitive performance. However, the choice of tokenizer can significantly impact the performance of n-gram models, as shown by Jimenez et al., (2018). Wang et al., (2023) further emphasized the importance of proper pre-training methods for visual tokenizers in multimodal large language models. These studies collectively highlight the need for careful consideration of tokenization methods in language model development.

Alyafeai and Ahmad, (2021) and Alkaoud and Syed, (2020) have explored the importance of tokenization in Arabic NLP, with Alyafeai introducing new tokenization algorithms and Alkaoud proposed strategies that consider Arabic's complex morphology. Toraman et al., (2022) further explore the impact of tokenization on language models, finding that morphological and word-level tokenizers can outperform standard methods in certain cases. These findings are particularly relevant for large language models, as they suggest that tokenization can significantly affect model performance.

To the best of our knowledge, there has been limited understanding of how morphological-based tokenizers affect the learning rate of models. Additionally, there has been limited research on how morphological-based tokenizers interplay with the vocabulary size of the models and whether they require fewer or more tokens to perform well.

The main aim of this study is to answer the following research questions:

> RQ1) What is the effect of different tokenization methods on the performance of downstream tasks using an Arabic language model? Specifically, does morphological-level (e.g.: Farasa) tokenization provide benefits compared to commonly used approaches like WordPiece and BPE? How would a morphological-based approach be generalized for different Arabic dialects?
>
> RQ2) How does model performance vary when tuning the vocabulary size, and how does this relationship differ across tokenization methods? There exists a trade-off between model size/parameters and performance based on vocabulary size; however, the nature of this relationship may depend on the tokenization approach. This was explored in this study.

The paper is organized with the following sections. Related work is discussed in section 2, the experiment is explained in section 3, results and discussions are all furnished in section 4 and finally conclusions are discussed in section 5.

## 2. Related Work

In large language models, tokenization is known as the division of text into a list of tokens. These tokens are designed to be feed to the architecture of large language models which is a crucial initial step in Natural Language Processing (NLP) systems (Metke-Jimenez et al., 2011). There are several approaches used to tokenize text. In contemporary literature, Byte Pair Encoding (BPE) (Sennrich et al., 2016b) and WordPiece (Schuster and Nakajima, 2012) have gained prominence as prevalent tokenization algorithms in language model pretraining research. BPE is criticized for its suboptimality in language pretraining (Bostrom and Durrett, 2020). A comparison with WordPiece and Word-level tokenization by (Nayak et al., 2020) reveals limitations in vocabulary space utilization (Gong et al., 2018). To address BPE's shortcomings, studies explore subword regularization for neural machine translation (Kudo, 2018). SentencePiece, implementing BPE and Unigram language models, is introduced as an alternative tokenization method (Kudo, 2018).

Morphological analysis emerges as a promising approach to enhance training efficiency and downstream performance (Park et al., 2020). Rule-based tokenization algorithms, incorporating lexicons and semantic parsing, extend methods to cross-lingual settings (Vasiu and Potolea, 2020). Hybrid tokenization approaches that combine coarse and fine-grained representations are explored (Hiraoka et al., 2021). Multi-grained tokenization methods, capturing multi-word representations, are incorporated into models, introducing increased computational complexity (Zhang et al., 2021). An emerging line of research focuses on enabling gradient-based learnable representations in the tokenization step of the pipeline (Tay et al., 2022)

Research on tokenization for low-resource languages, such as Thai (Lowphansirikul et al., 2021), Turkish (Toraman et al., 2022), Tibetan-to-Chinese machine translation (Zhang et al., 2021), and Part-of-Speech Tagging (Ding et al., 2019), highlights the importance of tailored approaches. Morphological-level tokenizers have shown promise in improving performance (Toraman et al., 2022). Proposals for Kurdish tokenization using morphological analysis (Alfaidi et al., 2023), and the exploration of morphological features in Korean language tokenization (Park et al., 2021) contribute valuable insights. Additionally, several investigations were performed on the impact of tokenziers on Arabic language models (Alyafeai et al., 2023)

In Arabic Natural Language Processing (NLP), subword tokenization is widely adopted, exemplified by HULMonA's use of WordPiece and AraBERT's use of SentencePiece (Abdelali et al., 2021; Antoun et al., 2020; ElJundi et al., 2019). These models demonstrate the effectiveness of subword tokenization, particularly when integrated with FARASA morphological segmentation.

The tokenisation process is considered a sensitive problem in comparative studies assessing the tokenization's impact on machine translation and BERT model fine-tuning; especially in in Arabic text processing (Oudah et al., 2019). Specific studies focused on the choice of tokenizer for n-gram models (Jimenez et al., 2018) and proper pre-training methods for visual tokenizers in multimodal large language models (Wang et al., 2023) reinforce the need for careful consideration of tokenization methods in language model development.

Noteworthy studies by (Alyafeai and Ahmad, 2021) introduce new tokenization algorithms, while (Alkaoud and Syed, 2020) propose strategies considering the complex morphology of the Arabic language. Additionally, (Toraman et al., 2022) explore the impact of tokenization on language models, finding that morphological and word-level tokenizers can outperform standard methods in certain cases. Although their study covered the Turkish language, their experiment was still relevant for Arabic since both languages have a very rich morphological structure.

## 3. Experiment

This paper proposes a pipeline that generates a total of twelve models. It consisted of 5 main steps: 1) Arabic corpus collection, 2) data text-processing, 3) tokenizer training, 4) model pre-training and finally 5) model fine-tuning.

Arabic corpus collection has adopted the OSCAR Arabic sub-set (Abadji et al., 2022). It has been utilized to ensure the rapid development and consistent reproduction of results. The second step was text pre-processing. The paper has developed a specific pre-processor algorithm that leveraged precious work made with the Arabert Preprocessor algorithm (Antoun et al., 2020).

The dataset preprocessor was designed specifically for the OSCAR Arabic subset. The pre-processing steps involved the removal of non-Arabic sentences. Moreover, multiple specialized filtering methods were implemented to refine the dataset further. These methods included the removal of documents that were extracted from navigation pages, short paragraphs, and other extraneous elements that might introduce noise or distort the underlying linguistic patterns. The second step was to remove any unnecessary or noisy text in each document, such as the removal of HTML makeups, non-digit repetitions, alphanumeric and the replacement of URLs, mentions, and emails with placeholders. More Arabic-specific filtering was used such as the removal of tatweel (e.g., a type of justification used in some cursive scripts), and diacritics (i.e., tashkeel), and the replacement of Hindi digits with Arabic ones. This was done in line with AraBert study (Antoun et al., 2020).

The next step of the pipeline was the tokenizers' training. This study has used a total of four types of tokenizers, namely Byte Pair Encoding (BPE), WordPiece, BPE with Farasa pre-processor (Morphological-level) and Word-level tokenizers. Each one of them requires a specific vocabulary size as an essential parameter for its training. This study has tested a total of three different vocabulary sizes to examine the effect of this variable on the performance of the tokenizers. The chosen vocabulary sizes were 16k, 28k and 44k. The choice of these numbers was based on the choices made by Toraman et al., (2022) and based on various tests done during the initiation of this study. This step has produced a total of 12 different tokenizers based on the tokenizer's type and its vocabulary size. HuggingFace tokenizers library was used to train the tokenizers (Dhyani, 2021).

It's worth noting that the dataset was segmented with Farasa library for the BPE-with-Farasa model training and fine-tuning. The rest of the tokenizers received the text with the aforementioned text preprocessing and without the Farasa pre-processing.

Table 1: Configuration Details of the Safaya Model.

| Parameters  | 48M       |
|-------------|-----------|
| Train data  | 96GB      |
| Layers      | 8         |
| Heads       | 8         |
| Hidden size | 512       |
| Batch size  | 128       |
| Max length  | 512 token |
| Train time  | n/a       |
| Hardware    | TPU v3-8  |

Below is a brief explanation of each tokenization method used in this study:
- **Byte Pair Encoding (BPE)** serves as a commonly employed tokenizer in the realm of pre-trained language models (Sennrich et al., 2016a). Operating at a granularity midway between characters and words, BPE generates tokens primarily as subwords, contingent on vocabulary size.
- **Morphological tokenizer (Farasa) with BPE** analysis takes a different approach than the aforementioned tokenizers. Morphological tokenizer splits words into suffixes and word stems that carry semantic depth, surpassing tokens obtained through overlapping frequency or likelihood. In this context, we explored the use of Farasa (Abdelali et al., 2016). The Farasa algorithm has proved to be an effective tokenizer for the Arabic language (Antoun et al., 2020). While morphological-level tokenization captures grammatically interpretable character sequences and facilitates learning semantics based on word suffixes, a drawback lies in the non-splitting of word stems, contributing to a sizable set that necessitates inclusion in the vocabulary, in addition to its inability to handle new words. Therefore, BPE was used additionally to mitigate the aforementioned issues.
- **WordPiece**, akin to BPE, revolves around the amalgamation of characters within documents (Wu et al., 2016). Its point of departure from BPE lies in its objective to maximize the likelihood score of language modelling. Both WordPiece and BPE operate on frequency-based algorithms, aiming to enhance the modelling prowess of individual tokens. This enables them to tokenize words not encountered during the training of the tokenizer.
- **Word-level tokenization** operates at the surface forms of words, effectively segmenting text based on the spaces between words. Unlike other methods, word-level tokenization requires no dedicated vocabulary training, as it can be implemented by simply splitting text using whitespace characters. However, a noteworthy drawback is the increased vocabulary size needed to adequately tokenize the same amount of text compared to alternative methods (Rai and Borah, 2021). The finite size of the vocabulary in language modelling introduces the possibility of encountering frequent out-of-vocabulary or unknown tokens using this approach.

Table 2 : Parameter Settings for BERT Model Training on Downstream Tasks used for training the model.

|  | News Classification | Hate Speech Detection | Sentiment Analysis | Natural Language Inference |
|---|---|---|---|---|
| Epochs | 10 | 5 | 10 | 3 |
| Max length | 512 | 512 | 512 | 512 |
| Batch size | 50 | 50 | 50 | 50 |
| Learning rate | 1e-5 | 1e-5 | 1e-5 | 1e-5 |
| Train size | 36400 | 4676 | 58891 | 20000 |
| Test size | 9100 | 1170 | 14723 | 5000 |

The fourth step of the pipeline is model pre-training. This study has used the Safaya Bert model (Safaya et al., 2020), see table 2. This model was initially trained on the OSCAR dataset (Abadji et al., 2022) and chosen due to its relatively medium size which allows for fast training while keeping the results of this study relevant to larger models. The approach of using an existing model while replacing its tokenizer reflects on the implications of replacing the tokenizers of existing models with tokenizers that were retrained on Arabic datasets or extending the existing tokenizers with a larger vocabulary size to support the Arabic language.

The initial configuration of the model is laid out in Table 2. The model matches the RoBERTa pretraining procedure and configuration following the architecture of BERT-Medium in terms of the number of layers, attention heads, and hidden size.

Employing the AdamW optimizer by (Loshchilov and Hutter, 2019), our configuration specifies $\beta_1$ as 0.90, $\beta_2$ as 0.98, and $\epsilon$ as 1e-6. The training regimen involves linear scheduling with a warmup ratio of 1e-2, reaching a peak learning rate of 5e-5, and gradient accumulation over 22 steps. Additionally, the remaining hyperparameters align with the RoBERTa configuration proposed by (Liu et al., 2019).

It's worth noting that the re-training of the model was kept computationally simple to avoid overtraining, which goes in hand with similar studies (Toraman et al., 2022). This phenomenon was observed by Xue et al., (2021) and proved that overtraining a model could influence high scores when models are retrained with very simple tokenizers such as character-based tokenizers. Therefore, the re-training aim was not to overfit the model but to acquaint it with the new tokenizer.

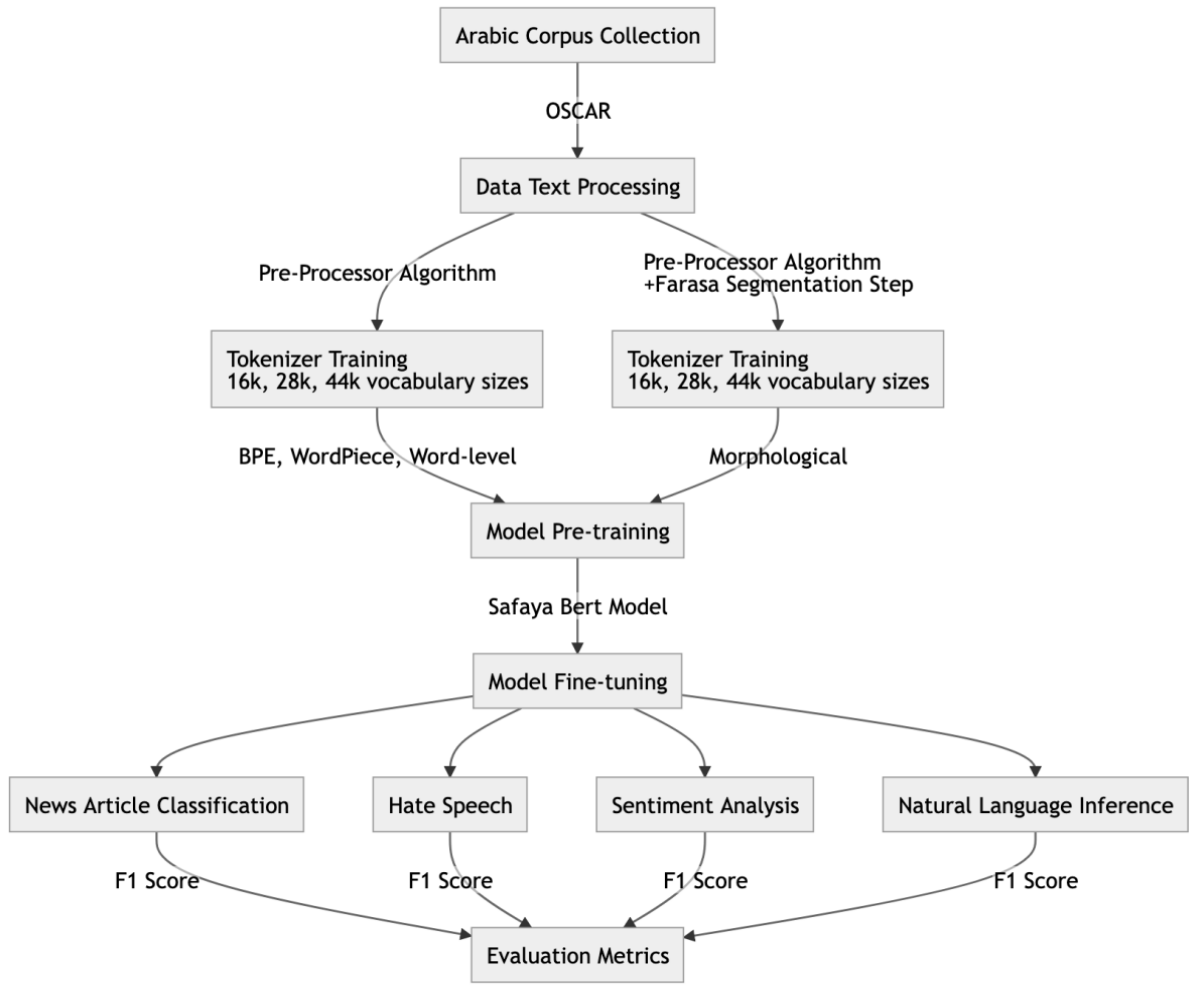

Figure 1: A simple illustration of the choices made in the experiment. The study used the safaya-model, retrained it on three different vocabulary sizes and each model was then fine-tuned on four different downstreaming tasks. This resulted in the fine-tuning of a total of 12 different models.

This experiment design has resulted on the training of 12 models since the study used a single model, four tokenizers and three vocabulary sizes as explained earlier. Please see Figure 2 for more details [1].

Finally, the final step of the tokenizer was fine-tuning. The fine-tuning step involved the fine-tuning of each of the 12 retrained models on 4 different downstream tasks. This step aimed to measure the performance of the re-trained model on different downstream tasks and how the new tokenizers could have impacted its performance.

Four downstream tasks were used and explained in the following points:

- **News Article Classification**: This task involves the classification of news articles into predefined categories, constituting a text sequence classification challenge. The utilized SANAD dataset, sourced from prominent news websites spanning diverse categories such as Sports, Finance, and Culture, forms the foundation for this investigation (Einea et al., 2019).

---

[1] Models are released under https://huggingface.co/nourmorsy

- **Hate Speech Detection**: This task aims to detect the presence of hate speech within a given text sequence, particularly targeting individuals or communities with varying backgrounds. In this research, we leverage the Arabic Levantine Hate Speech dataset, encompassing political tweets labelled as normal, abusive, or exhibiting hate speech emanating from intense political debates across various Arabic nations (Mulki et al., 2019).
- **Sentiment Analysis**: Sentiment Analysis entails the classification of text sequences to ascertain the author's emotional disposition. Our investigation draws upon the ASAD dataset, which encompasses a broad spectrum of tweets expressing positive, negative, and neutral sentiments, composed in various Arabic dialects, including the Egyptian variant (Alharbi et al., 2020).
- **Natural Language Inference**: Natural Language Inference involves predicting whether the second sentence can be inferred from the first when presented with two sentences. The dataset, developed by Laurer et al.,(2023), employed in this task incorporates three types of semantic relations: entailment (where the first sentence entails the second), neutral (indicating no apparent relation between the sentences), and contradiction (where the first sentence contradicts the second). The NLI dataset, inclusive of Arabic, spans a multitude of languages and was curated employing state-of-the-art open-source machine translation models, the Arabic subset was used during the training of this study (Laurer et al., 2023).

Each one of the downstream tasks was measured against the F1 score to identify its performance for each downstreaming task. The fine-tuning was done to make it computationally simple to reduce the overfitting and isolate the effect of the tokenizers on the performance of the re-trained model.

## 4. Results and Discussion

In our investigation, BPE with Farasa consistently outperformed other tokenization strategies in three critical tasks: News Classification, Hate Speech Detection, and Sentiment Analysis. The morphological nature of Farasa, when combined with BPE, facilitated a superior organization of token embeddings. The tasks benefitted from this approach, likely due to the model's ability to capture the complex morphological structures of Arabic, providing an enhanced representation which improved performance, see table 3.

In contrast, our analysis revealed challenges for BPE with Farasa in sentiment analysis, especially when dealing with datasets predominantly in dialects. Specifically, the Egyptian dialect posed segmentation difficulties for Farasa, leading to suboptimal performance. Additionally, the limited availability of dialect text in the training dataset impacted the overall efficiency of BPE with Farasa in sentiment analysis, making it crucial to address these challenges for improved model outcomes.

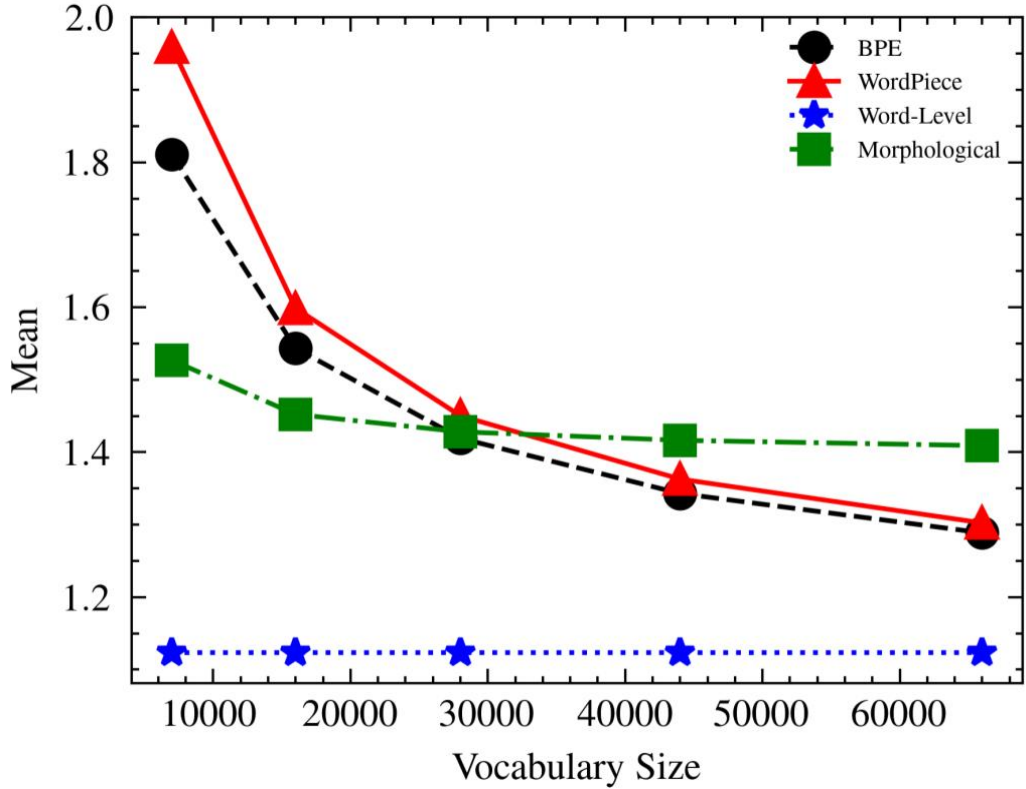

Figure 3: The graph displays the Tokenizers' performance based on token-to-word ratio across various vocabulary Sizes. The utilization of morphological tokenizers, specifically Farasa and BPE, has effectively stabilized the token-to-word ratio, mitigating the substantial reduction introduced by BPE and WordPiece. While this stabilization aids in vocabulary size reduction, our findings highlight a nuanced trade-off: diminishing the vocabulary size without incorporating morphological considerations results in a discernible decline in the training performance of Arabic language models.

BPE with Farasa exhibited limitations in generalization, particularly in handling out-of-vocabulary (OOV) words. For example, words such as ('يتحدثها' means he speaks it) could be easily segmented using Farasa into the following segments ('يتحدث+ها ' means 'he speaks' and 'it'). However, words like ('مبيتحدثهاش' means 'he didn't speak it' in Egyptian dialect) were not segmented properly using Farasa; which is a known limit (Alfaidi et al., 2023). This hinders the model's ability to handle the aforementioned limitations as an out-of-vocabulary, limiting its ability to perform efficiently.

While our study findings contradict (Toraman et al., 2022) since their recommendation that BPE and WordPiece performed best in the Turkish language, Toraman et al., (2022) concluded that morphological-based tokenizers inherit the limitations of their morphological segmenter which was present in this study as well; leaving out-of-vocabulary words as the main issue found with Farasa and their morphological tokenizer too. To alleviate that in the Arabic language, recent research by (Alkaoud and Syed, 2020). Their paper proposed an approach for an accurate representation of OOV words, but it was tailored for Word2Vec models. However, integrating Alkaoud's method with modern LLM architectures would require additional development. Addressing this limitation is vital for enhancing the robustness of BPE with Farasa, ensuring a more comprehensive understanding of the Arabic language.

Our study underscores the importance of a balanced dataset that incorporates both dialect and formal Arabic for downstream tasks. This aligns with a similar study that advocated for multiple monolingual tokenizers to enable the models to learn multiple dialects efficiently which aligned with our findings (Rust et al., 2021). However, whether employing different tokenizers for each Arabic dialect surpasses the benefits of replacing Farasa with a more dialect-independent segmenter remains an unexplored area, warranting further investigation in future research.

Table 3: Performance Metrics of the Asafaya Model on Downstream Tasks. Precision (P), Recall (R), and F1 scores are reported for News Classification, Hate Speech Detection, Sentiment Analysis, and Natural Language Inference across different tokenization approaches and model sizes (16k, 28k, 44k). BPE (Byte Pair Encoding), WP (WordPiece), BPE_Farasa (Farasa using Byte Pair Encoding), and Word (WordLevel) represent the different tokenization strategies, each with distinct performance characteristics.

|  |  | News Classification | | | Hate Speech Detection | | | Sentiment Analysis | | | Natural Language Inference | | |
| --- | --- | --- | --- | --- | --- | --- | --- | --- | --- | --- | --- | --- | --- |
|  |  | P | R | F1 | P | R | F1 | P | R | F1 | P | R | F1 |
| 16K | BPE | 0.975 | 0.974 | 0.974 | **0.662** | **0.733** | **0.694** | **0.826** | **0.831** | **0.827** | 0.533 | 0.525 | 0.522 |
|  | WP | 0.972 | 0.972 | 0.972 | 0.624 | 0.697 | 0.652 | 0.783 | 0.788 | 0.785 | 0.523 | 0.519 | 0.516 |
|  | BPE Farasa | **0.975** | **0.974** | **0.974** | 0.691 | 0.717 | 0.679 | 0.791 | 0.793 | 0.792 | **0.568** | **0.569** | **0.568** |
|  | Word | 0.974 | 0.973 | 0.973 | 0.633 | 0.700 | 0.665 | 0.741 | 0.755 | 0.746 | 0.548 | 0.542 | 0.541 |
| 28K | BPE | 0.976 | 0.976 | 0.976 | 0.643 | 0.714 | 0.673 | **0.816** | **0.820** | **0.817** | 0.538 | 0.532 | 0.528 |
|  | WP | 0.974 | 0.974 | 0.974 | 0.641 | 0.712 | 0.671 | 0.769 | 0.775 | 0.771 | 0.543 | 0.535 | 0.533 |
|  | BPE Farasa | **0.977** | **0.976** | **0.976** | **0.652** | **0.723** | **0.684** | 0.792 | 0.789 | 0.790 | **0.578** | **0.573** | **0.572** |
|  | Word | 0.972 | 0.972 | 0.972 | 0.642 | 0.712 | 0.673 | 0.744 | 0.755 | 0.748 | 0.547 | 0.542 | 0.541 |
| 44K | BPE | 0.974 | 0.974 | 0.974 | 0.634 | 0.705 | 0.663 | **0.824** | **0.827** | **0.825** | 0.539 | 0.533 | 0.531 |
|  | WP | 0.973 | 0.973 | 0.973 | 0.627 | 0.697 | 0.654 | 0.773 | 0.780 | 0.776 | 0.534 | 0.529 | 0.527 |
|  | BPE Farasa | **0.978** | **0.978** | **0.978** | **0.740** | **0.723** | **0.684** | 0.788 | 0.790 | 0.789 | **0.567** | **0.564** | **0.563** |
|  | Word | 0.974 | 0.974 | 0.974 | 0.644 | 0.712 | 0.675 | 0.759 | 0.766 | 0.761 | 0.547 | 0.542 | 0.540 |

**Impact of Vocabulary Size on Model Performance**

Changes in vocabulary size had a minimal impact on the F1 scores of models across various downstream tasks. Unlike studies that adjusted model size based on vocabulary size, our approach maintained the same model size. Notably, the performance of model sizes with vocabulary sizes of 16k closely resembled those with 28k and 44k, challenging the findings of (Toraman et al., 2022). This indicates that changing the vocabulary size without adjusting the model size does not significantly influence model performance, necessitating further research in this domain.

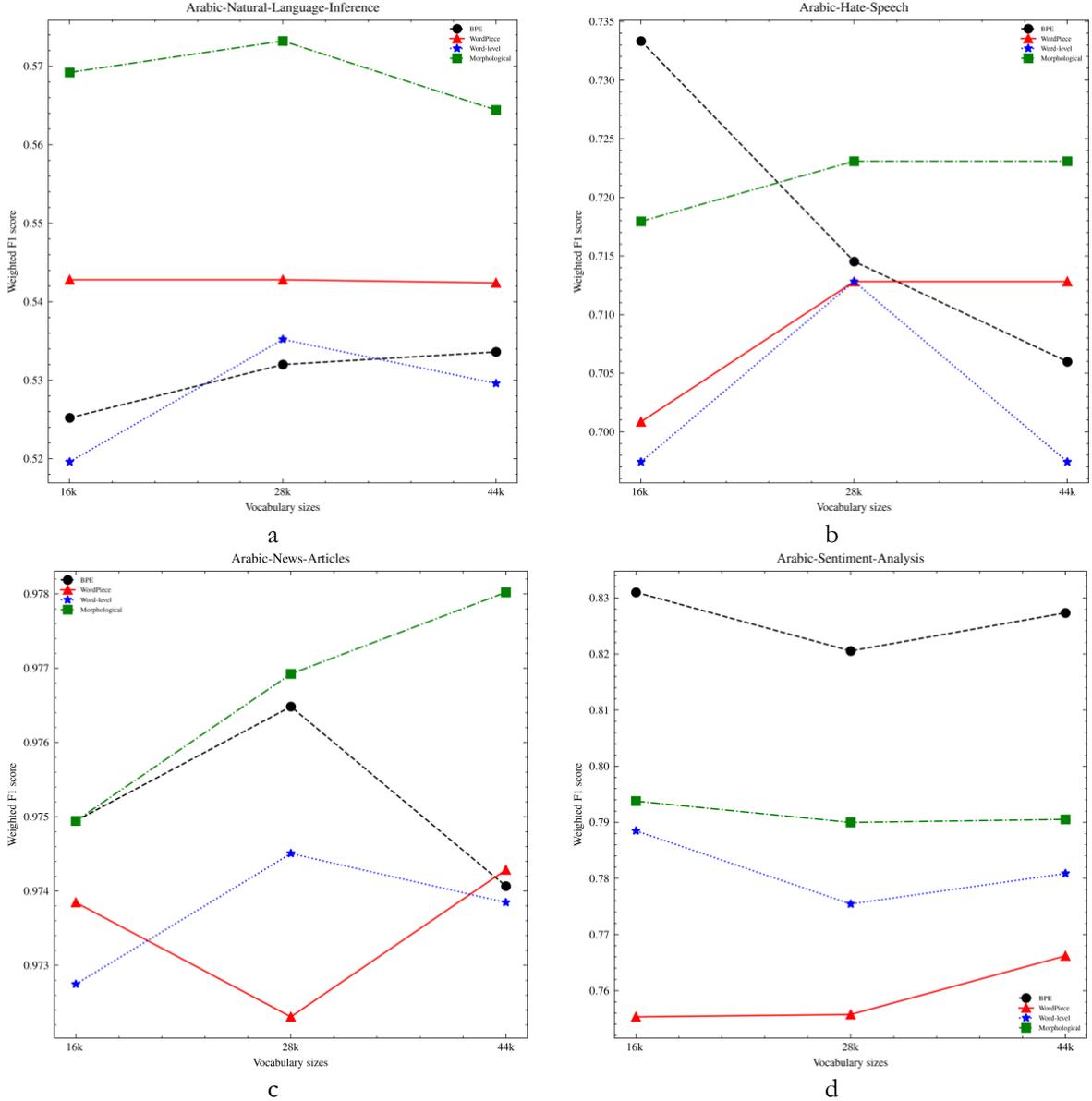

Figure 4: display of the f1 performance of the trained models on different downstream tasks and how f1 changes according to vocabulary size.

Contrary to Toraman et al., (2022), our study revealed that changing the vocabulary size had limited effects on model performance, consistent with (Alyafeai et al., 2023). The negligible changes among vocabulary sizes, particularly in F1 scores, suggest that modifications in tokenization and vocabulary size may not substantially impact the overall model performance in our experimental setting, see Table 1.

However, our findings are different of Toraman et al., (2022) because we maintained a consistent model size across varying vocabulary sizes whilst they changed the model size to fit the expanding vocabulary size. Therefore, our study affirms that changing the vocabulary size does not influence model performance unless accompanied by alterations in model size. Additional research is imperative to delve deeper into this relationship and understand the interplay between vocabulary size and model parameters in the context of Arabic language models.

In our comparative analysis of model sizes, we observed intriguing patterns in the performance of models with varying tokenization strategies and vocabulary sizes. Despite maintaining a consistent model size across different vocabulary sizes (16k, 28k, and 44k), the performance variations were not as pronounced as anticipated. This contrasts with traditional approaches that adjust model sizes based on vocabulary sizes to enhance learning capacity. The limited impact of vocabulary size on model performance prompts a deeper exploration into the relationship between tokenization, vocabulary size, and model architecture. Future studies should systematically investigate the potential benefits and trade-offs associated with adjusting model sizes alongside vocabulary changes to gain a more nuanced understanding of their interconnected dynamics.

Leveraging pre-trained models, especially those trained on large-scale corpora, could potentially mitigate some of the challenges encountered in our study. These models encapsulate extensive linguistic knowledge and can serve as a strong foundation for downstream tasks. Exploring transfer learning methods, such as fine-tuning pre-trained models on task-specific datasets, could further optimize model performance (Dhyani, 2021). Additionally, the adaptation of multi-task learning strategies, where models are trained simultaneously on multiple related tasks, might yield more robust representations (Aribandi et al., 2022). Future research endeavors should delve into the integration of transfer learning techniques to amplify the capabilities of Arabic language models, addressing both generalization challenges and task-specific nuances, especially the transfer learning from English language to Arabic (Tran, 2020).

**BPE-Farasa Advantage and Recommendations**

Overall, BPE with Farasa emerged as a highly effective tokenization strategy for Arabic language understanding tasks, particularly excelling in semantic-centric tasks like Natural Language Inference and News Article Classification. The integration of Farasa's morphological analysis stabilized BPE's performance, showcasing a competitive performance at a smaller vocabulary size (16k). However, this study's results highlighted the limitations of the Farasa segmenter on different Arabic dialects and how that increases out-of-vocabulary words causing the underlying model to behave inefficiently. Further work should ensure the use of a dialect-independent segmenter or the use of multiple mono-dialect tokenizers to ensure the high efficiency possible in terms of tokenization as suggested by other studies (Rust et al., 2021). Also, our results emphasize the need for further exploration into adaptive tokenizers based on downstream applications, tailoring approaches to specific tasks and linguistic nuances in future research endeavors (Vasiu and Potolea, 2020).

**Model Performance in Natural Language Inference (NLI) Tasks Across Tokenization Strategies**

Natural Language Inference (NLI) tasks, depicted in Figure 5-a, shed light on the dynamics of tokenization strategies. BPE and WordLevel exhibited comparable performances, both showcasing improvements with an expanded vocabulary size. Notably, Farasa's BPE outperformed others, securing the highest average F1 score of 0.563 across all vocabulary sizes. WordPiece demonstrated relatively stable performance, with an average F1 of 0.54. This suggests that morphological tokenization, especially when combined with BPE, contributes significantly to capturing semantic relationships in NLI tasks. The distinct advantages of Farasa's BPE in this context emphasize the need to tailor tokenization strategies based on the specific linguistic intricacies inherent in different natural language processing tasks.

**Comparative Analysis of Hate Speech Detection Across Vocabulary Sizes**

Hate Speech Detection, as illustrated in Figure 6-b, presented a unique set of challenges and successes across different vocabulary sizes. BPE-Farasa excelled in larger vocabularies, achieving an F1-score of 0.678, closely followed by WordPiece at 0.675. However, BPE demonstrated superior performance in smaller vocabulary sizes, experiencing a decline in larger sizes. This underscores the sensitivity of hate speech detection to tokenization choices and the impact of vocabulary size on model outcomes.

**Implications of Tokenization Choices on News Article Classification**

News article classification, as observed in Figure 7-c, revealed intriguing implications of tokenization choices. BPE-Farasa emerged as the top performer, particularly in larger vocabularies, boasting an exceptional F1-score of 0.978. This underscores the significance of morphological analysis in enhancing the model's understanding of formal Arabic in news articles. The findings align with the importance of morphological depth of the Arabic language as captured by BPE-Farasa. The implications for news article classification underscore the task-specific advantages of certain tokenization strategies, urging researchers to consider these factors when designing models for news-related applications.

**Sentiment Analysis and Tokenization Strategy Dynamics**

Sentiment analysis, depicted in Figure 8-d, showcased the dynamics in tokenization strategy performances. Surprisingly, BPE achieved the highest average F1 of 0.83, even with a decrease in performance as vocabulary size expanded. This suggests a unique reliance on surface-level patterns in sentiment analysis, where BPE excels despite potential challenges with out-of-vocabulary words. Notably, BPE-Farasa and WordLevel closely followed with average F1-scores of 0.78. WordPiece lagged with the least efficient performance at 0.73.

**Computational Efficiency and Training Dynamics**

Examining the computational efficiency and training dynamics of the models is crucial for practical deployment. Despite the variations in tokenization strategies and vocabulary sizes, our study observed comparable training times across different models. The computational efficiency of BPE with Farasa was notable, demonstrating stable training dynamics. This efficiency could be attributed to Farasa's morphological analysis, enabling a more streamlined learning process. However, further investigation into the computational demands of larger datasets and model sizes is warranted to ensure scalability and efficiency in real-world applications.

Ensuring model robustness and generalization across diverse domains is imperative for real-world applicability. While BPE with Farasa exhibited superior performance in specific tasks, its robustness across a wide range of domains requires scrutiny. Our study highlights the need for models to adapt seamlessly to the dynamics of Arabic whether its modern standard of Arabic dialect. Assessing the model's capacity to generalize to unseen data and domains is essential, and future research should focus on enhancing the adaptability of tokenization strategies to ensure reliable performance across different linguistic landscapes.

**Ethical Considerations and Bias in Model Performance**

Ethical considerations and the potential bias embedded in model performance are paramount concerns in natural language processing. Our study acknowledges the importance of addressing biases, especially in Arabic dialects where cultural nuances play a significant role.

Moreover, the influence of training datasets on bias is a critical aspect that demands attention. The limited representation of dialects in our training dataset may have implications for bias in model predictions, particularly in tasks related to sentiment analysis and hate speech detection. Future studies should prioritize building diverse datasets covering all Arabic dialects in addition to MSA to mitigate such an issue.

**Future Directions and Research Recommendations**

Future research can focus on several points to extend the literature's understanding of the complexity of Arabic language and its impact on LLM architecture. Firstly, investigating the effectiveness of alternative tokenization strategies, beyond the explored BPE and WordPiece, holds promise. The development and integration of novel tokenization methods that cater specifically to the intricacies of Arabic morphology may unlock further improvements. Additionally, expanding datasets to encompass a more diverse range of dialects and linguistic variations can enhance the generalizability of models. Furthermore, exploring the integration of domain-specific knowledge and ontologies could refine the contextual understanding of Arabic language models.

Moreover, researchers should delve into refining existing segmentation tools, such as Farasa, to overcome challenges associated with dialects and out-of-vocabulary words. Evaluating the adaptability of such tools across different tasks and domains is crucial for ensuring comprehensive language coverage. As we move forward, collaborative efforts among researchers in the Arabic natural language processing community can foster the development of standardized benchmarks and datasets, facilitating fair comparisons and advancements in the field.

Finally, the study has focused on token embedding, while neglecting some crucial tasks as Name Entity Recognition and Part of Speech tagging. This could potentially identify further challenges arising from the complexity of such tasks on sentence structure which was not covered using the chosen tasks in this paper.

## 5. Conclusion

In conclusion, our comprehensive exploration into the impact of tokenization strategies and vocabulary sizes on the performance of Arabic language models sheds light on critical aspects that shape the landscape of natural language processing in this linguistic domain. Through rigorous experimentation and analysis, several key findings have emerged, guiding our understanding of the nuances inherent in training models for downstream tasks.

The superior performance of BPE with Farasa in tasks such as News Classification, Hate Speech Detection, and Sentiment Analysis underscores the significance of leveraging morphological analysis to capture the intricacies of the Arabic language. However, challenges surfaced, particularly in sentiment analysis, where dialect-specific segmentation issues posed hurdles for Farasa. These findings emphasize the need for continuous refinement of tokenization strategies to navigate the diverse linguistic terrain of Arabic.

Our study also explored the computational efficiency, revealing the stable training dynamics of BPE with Farasa, suggesting its viability for real-world applications. The limited impact of vocabulary size on model performance challenges existing assumptions, indicating the need for a nuanced understanding of the interplay between vocabulary, model size, and downstream tasks.

Robustness and generalization across domains emerged as critical considerations, urging further research to ensure the adaptability of models to diverse linguistic contexts. Ethical dimensions, especially in mitigating bias and promoting fairness in model predictions, demand ongoing attention, as our study highlights potential challenges in sentiment analysis due to dialect-specific biases.

As the field of Arabic language processing continues to evolve, our work contributes valuable insights and sets the stage for future investigations. The recommendations for future research include addressing the challenges in sentiment analysis, refining tokenization strategies for improved robustness, and expanding datasets to encompass the rich diversity of Arabic linguistic nuances. By navigating these complexities, we pave the way for the responsible development and deployment of Arabic language models that meet the demands of real-world applications while upholding ethical standards and promoting fairness across diverse linguistic landscapes.